\documentclass[conference]{IEEEtran}
\IEEEoverridecommandlockouts
\usepackage{cite}
\usepackage{amsmath,amssymb,amsfonts}
\usepackage{algorithmic}
\usepackage{graphicx}
\usepackage{textcomp}
\usepackage{xcolor}
\usepackage{multirow}
\usepackage{subcaption}
\usepackage[a4paper, total={184mm,239mm}]{geometry}

\def\BibTeX{{\rm B\kern-.05em{\sc i\kern-.025em b}\kern-.08em
    T\kern-.1667em\lower.7ex\hbox{E}\kern-.125emX}}
\graphicspath{{./figures}}
\begin{document}
\title{
	Harden Deep Neural Networks Against Fault Injections Through Weight Scaling
	\thanks{
		This research is based on results obtained from a project, JPNP16007,
		commissioned by the New Energy and Industrial Technology Development
		Organization (NEDO).
	}
}

\makeatletter
\newcommand{\linebreakand}{\end{@IEEEauthorhalign} \hfill\mbox{}
	\par
	\mbox{}\hfill\begin{@IEEEauthorhalign}}
\author{
	\IEEEauthorblockN{1\textsuperscript{st} Ninnart Fuengfusin}
	\IEEEauthorblockA{\textit{Graduate School of Life Science and Systems} \\
		\textit{Kyushu Institute of Technology}\\
		Kitakyushu, Japan \\
		ninnart@brain.kyutech.ac.jp}
	\and

	\IEEEauthorblockN{2\textsuperscript{nd} Hakaru Tamukoh}
	\IEEEauthorblockA{\textit{Graduate School of Life Science and Systems} \\
		\textit{Research Center for Neuromorphic AI Hardware}\\
		\textit{Kyushu Institute of Technology}\\
		Kitakyushu, Japan\\
		tamukoh@brain.kyutech.ac.jp}
}
\makeatother
\maketitle

\begin{abstract}
	Deep neural networks (DNNs) have enabled smart applications on hardware devices.
	However, these hardware devices are vulnerable to unintended faults caused by aging,
	temperature variance, and write errors. These faults can cause bit-flips in DNN
	weights and significantly degrade the performance of DNNs. Thus, protection against
	these faults is crucial for the deployment of DNNs in critical applications.
	Previous works have proposed error correction codes based methods, however these
	methods often require high overheads in both memory and computation. In this paper,
	we propose a simple yet effective method to harden DNN weights by multiplying
	weights by constants before storing them to fault-prone medium. When used, these
	weights are divided back by the same constants to restore the original scale. Our
	method is based on the observation that errors from bit-flips have properties
	similar to additive noise, therefore by dividing by constants can reduce the
	absolute error from bit-flips. To demonstrate our method, we conduct experiments
	across four ImageNet 2012 pre-trained models along with three different data types:
	32-bit floating point, 16-bit floating point, and 8-bit fixed point. This method
	demonstrates that by only multiplying weights with constants, Top-1 Accuracy of
	8-bit fixed point ResNet50 is improved by 54.418 at bit-error rate of 0.0001.
\end{abstract}


\section{Introduction}
In recent years, deep neural networks (DNNs) have enabled applications such as image
recognition \cite{liu2021swin}, object detection \cite{wang2024yolov10}, and natural
language processing \cite{touvron2023llama}. These applications form the backbones of
many critical systems, including autonomous cars \cite{grigorescu2020survey}, healthcare
\cite{zheng2024large}, and more.

However, as DNNs are deployed to hardware devices, they become vulnerable to fault
injection, which can be caused by aging \cite{dixit2021silent}, temperature variance
\cite{cha2017defect}, and write errors \cite{nozaki2019recent}. These faults can
manifest as bit-flips to DNN weights and can easily degrade DNN performances
\cite{hong2019terminal}. Without any protections, faults occurring in DNNs used in
critical applications may lead to serious consequences.

To address this issue, several research directions have been proposed. One of the
research directions is error-correction code (ECC) based methods
\cite{fuengfusin2023efficient, lee2022value, guan2019place, traiola2023machine}, which
use ECC to encode redundancy in the form of additional bits to DNN parameters and enable
ECC to correct a certain number of bit-flips. However, these methods often require high
overheads in both memory to encode redundancy and computation to locate and correct
error bits. The works in this direction \cite{fuengfusin2023efficient, lee2022value,
guan2019place} explore weight parameter characteristics to reduce memory overheads;
however, the computational overheads still remain high.

In this paper, our work proposes a simple yet efficient method that requires only
element-wise multiplication before writing and element-wise division after reading
weights from fault-prone mediums. We demonstrate that these processes reduce the overall
absolute error caused by bit-flips. Furthermore, we propose a method that reduces the
overall number of divisions by dividing output logits instead of weights. We show that
our method can be applied across three data types: 32-bit floating point (FP32), 16-bit
floating point (FP16), and 8-bit fixed point (Q2.5, with 1-bit sign, 2-bit integer, and
5-bit fraction). An overview of our proposed method is illustrated in Fig.
\ref{fig:overview}.

\begin{figure}[htbp]
	\centerline{\includegraphics[scale=0.275]{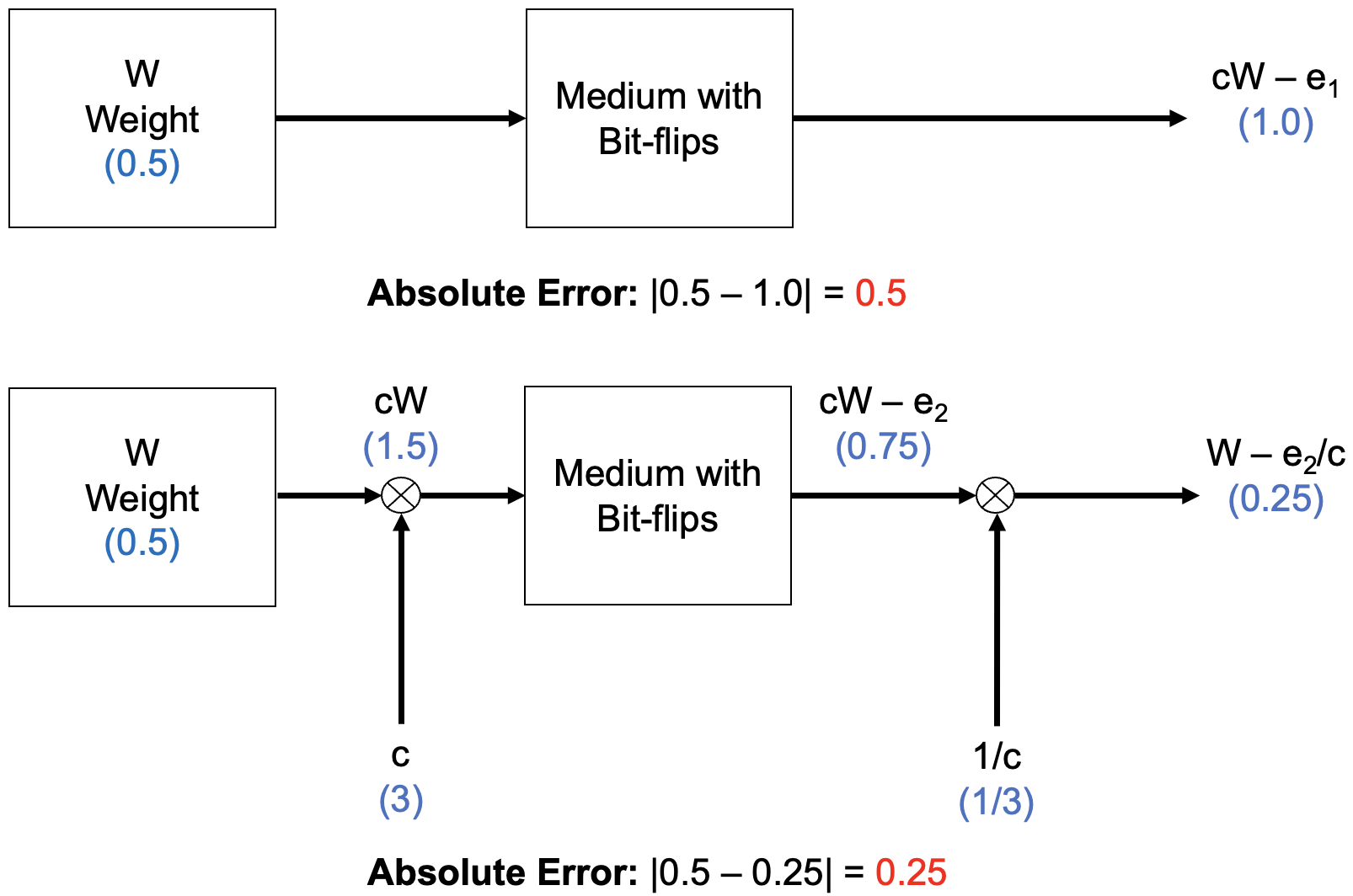}}
	\caption{
        Overview of our proposed method. \textbf{Top:} illustrates the baseline method,
        where a weight $W$ is directly written to fault-prone mediums. When read, this
        weight is affected by bit-flip errors in the form of $e_{1}$. \textbf{Bottom:}
        illustrates our proposed method, where a weight $W$ is multiplied by a constant
        $c$ before being written to fault-prone mediums. During deployment, the weights
        with faults $cW - e_{2}$ are read and rescaled back to the original scale $W -
        \frac{e_{2}}{c}$. By scaling and rescaling the weights, the overall absolute
        error caused by bit-flips is reduced. In this example, a bit-flip is injected
        into the 9-th bit position (1-th bit position is a sign bit) of the FP32 weight.
    }
	\label{fig:overview}
\end{figure}

Our main contributions are listed as follows:

\begin{itemize}
    \item This work demonstrates that simple element-wise multiplications and divisions
    can improve the robustness of DNNs against bit-flips.
    \item This work shows that our proposed method can be generalized across four
    ImageNet 2012 pre-trained models and three data types.
    \item This work proposes a method to reduce the number of divisions by dividing
    output logits instead of weights.
\end{itemize}

\section{Related Works}
This section discusses works related to hardening DNNs against fault injections. One
approach to protecting DNNs against bit-flips is to use ECC. ECC encodes redundancy in
the form of parity bits to DNN parameters, allowing it to correct a certain number of
bit-flips in the parameters. However, storing parity bits requires additional memory, so
ECC-based methods often focus on reducing memory overheads by exploiting weight
distribution characteristics.

Following this direction, in-place zero-space memory protection \cite{guan2019place} is
designed using ECC, specifically the extended Hamming code (64, 57), which can correct
up to one bit-flip error. This method also proposes a training strategy that penalizes
large weights, ensuring that large weights appear only at specific positions. This
approach ensures that most weights small, making the most significant bits (MSB)
after the sign bit likely to hold no information. As a result, this bit position can be
used to insert a parity bit.

In the same direction, value-aware parity insertion ECC \cite{lee2022value} is a method
based on ECC (64, 50), which can correct double bit-flips. This method is designed for
Q2.5 DNNs. To reduce memory overheads, it leverages the observation that the most weight
values are less than $\lvert 0.5 \rvert$, meaning the first two integer bits are likely
to hold no information. As a result, these bit positions can be used to insert parity
bits. If the weight values are greater than or equal to $\lvert 0.5 \rvert$, the last
two least significant bits (LSB) are used for parity bit insertion instead, ensuring
minimal information loss.

In another approach, instead of protecting all bit positions equally, efficient
repetition code for deep learning \cite{fuengfusin2023efficient} is designed to protect
only the bit positions close to the MSB, while avoiding protection for the bit positions
near the LSB. This approach is based on the observation that bit-flips in bit positions
near the MSB can cause dramatic changes compared to those near the LSB. Therefore, these
bit positions near the MSB are prioritized for protection.

ECC-based approaches are similar to our method, as both exploit the characteristics of
weight parameters and modify them either by inserting parity bits or scaling. A key
distinction is that our method incurs significantly lower computational overhead
compared to ECC-based methods. Our method only requires only element-wise multiplication
and division, whereas ECC-based methods involve encoding and decoding processes that can
be computationally expensive.

\section{Preliminaries}
In this section, we describe the prerequisites for our proposed method. Given a $i$-th
layer and $j$-th DNN weight denoted as $W_{i, j} \in \mathbb{R}$, its binary
representation is denoted as $B_{i, j} \in \{0, 1\}^{n}$, where $n$ is the number of
bits of the data type. Here, $B_{i, j, 1}$ denotes the MSB, and $B_{i, j, n}$ denotes
the LSB.

To model fault injections, a bit-flip at $B_{i, j, k}$ is defined as $M_{i, j, k} \in
\{0, 1\}$, where $M_{i, j, k} = 1$ indicates the presence of a bit-flip, while $M_{i, j,
k} = 0$ indicates no bit-flip. The probability of $M_{i, j, k}$ is defined in
\eqref{eq:bit-flip-prob}, where $BER$ is the bit-error rate, or probability of a
bit-flip.

\begin{equation}
	P(M_{i, j, k}) =
	\begin{cases}
		BER     & \text{if $M_{i, j, k} = 1$} \\
		1 - BER & \text{if $M_{i, j, k} = 0$}
	\end{cases}
	\label{eq:bit-flip-prob}
\end{equation}

To model bit-flip errors, let $f_{D}$ be a function that converts a binary
representation to its data type $D$ and let $b_{D}$ be a function that converts a data
type $D$ to its binary representation, where $D \in \{FP32,\, FP16,\, Q2.5\}$.
Therefore, the $i$-th layer and $j$-th weight with bit-flips, denoted as $\hat{W}_{i,
j}$, is defined in \eqref{eq:baseline}, where $M_{i, j}$ is a bit-flip mask representing
bit-flip occurrences for the weight at indices $i$ and $j$.

\begin{equation}
	\hat{W}_{i, j} = f_{D}(b_{D}(W_{i, j}) \oplus M_{i, j})
	\label{eq:baseline}
\end{equation}

Errors from bit-flip are founded using $e$ function defined in \eqref{eq:err}, where
$\oplus$ denotes as an element-wise XOR operation.

\begin{equation}
    \begin{split}
	    e(W_{i, j}, M_{i, j}) &= W_{i, j} - f_{D}(b_{D}(W_{i, j}) \oplus M_{i, j}) \\
        &= W_{i, j} - \hat{W}_{i, j}
    \end{split}
	\label{eq:err}
\end{equation}

\section{Proposed Methods}
After describing the prerequisites, we present our proposed method to harden DNNs
against fault injections through weight scaling in this section. Our method scales the
weights $W_{i}$ by $c_{i}$ before passing them to the $e$ function that injects
bit-flips into $c_{i}W_{i}$. After that, the weights are rescaled by dividing by the
same $c_{i}$.

\subsection{FP32 and FP16 Requirements}

To enable FP32 and FP16 models to operate with faults, it is necessary to avoid
bit-flips to the most significant bit of the exponent term or ensure $M_{i, j, 2} = 0$.
A bit-flip in this position can significantly alter the magnitude of the weight. For
example, $0.1$ can be changed to $\approx 3.403 \times 10^{37}$, and $1.0$ can be
changed to Infinity (\texttt{Inf}).

To enable this assumption, either ECC can be used to ensure data integrity, or
values can be clamped to a range of $(-2, 2)$, based on the observation that well-trained
weights are typically small values \cite{guan2019place, lee2022value}. From our
observations, only two weights from \texttt{torchvision} \cite{torchvision2016}
pre-trained weights—from AlexNet \cite{krizhevsky2012imagenet}, ResNet18
\cite{he2016deep}, ResNet50 \cite{he2016deep}, and DenseNet169 \cite{huang2017densely}
models—fall outside this range $(-2, 2)$. For example, clamping AlexNet weights to this
range results in a minor accuracy drop from 57.55 to 56.50. When all weights are within
this range, we can assume that the MSB of exponent term is always zero.

To enable this assumption in a lossless manner, a sparse matrix method based on
\cite{lee2022value} can be used to store the positions of values that fall outside the
range of $(-2, 2)$. Since the majority of weight values are within that range, the
overhead of this method should be minimal. By ensuring that the sparse matrix remains
lossless, weights outside the $(-2, 2)$ can be re-incorporated into the model later.

Enabling this assumption makes FP32 and FP16 models more robust compared to Q2.5 models.
While this assumption may introduce additional overhead, it is important to note that
our method with Q2.5 does not require it.

\subsection{Scaling, Rescaling, and Their Effects}
Before deployment to fault-prone mediums, the $i$-th layer weight $W_{i}$ is scaled by a
constant $c_{i}$, then written to the fault-prone mediums, as shown in
\eqref{eq:scaled-weight}. The weights with faults, based on our proposed method, are
denoted as $\tilde{W}_{i, j}$.

\begin{equation}
     c_{i}\tilde{W}_{i, j} = c_{i}W_{i, j} - e(c_{i}W_{i, j}, M_{i, j})
	\label{eq:scaled-weight}
\end{equation}

After receiving the scaled weight with fault injections, it can be recovered by dividing
by $c_{i}$, as shown in \eqref{eq:rescaled-weight},

\begin{equation}
	\tilde{W}_{i, j} = W_{i, j} - \frac{e(c_{i} W_{i, j}, M_{i, j})}{c_{i}}
	\label{eq:rescaled-weight}
\end{equation}

From our observations, the $e$ function is not scale-invariant, meaning that
$e(c_{i}W_{i, j}, M_{i, j}) \neq c_{i}e(W_{i, j}, M_{i, j})$. When $c_{i} > 1$, after
scaling $W_{i, j}$ with $c_{i}$, we found that, on average, bit-flips to scaled weights
cause higher absolute error from bit-flips: $\lvert e(c_{i}W_{i, j}, M_{i, j}) \rvert >
\lvert e(W_{i, j}, M_{i, j}) \rvert$. However, after rescaling back to the original
scale, the absolute error from bit-flips is reduced: $\lvert \frac{e(c_{i} W_{i, j},
M_{i, j})}{c_{i}} \rvert < \lvert e(W_{i, j}, M_{i, j}) \rvert$. These observations are
based on the following analysis.

\subsection{Analysis of the Effects of Scaling and Rescaling}
In this analysis, pseudo-weights were generated from -0.5 to 0.5 with an increment of
$0.01$. This range is chosen based on \cite{lee2022value}, which indicates that the most
of pre-trained weights from certain models fall within this interval.

Monte Carlo simulations were conducted for $10^{6}$ rounds by randomly injecting
bit-flips into pseudo-weights with a $BER = 10^{-1}$. This analysis was performed across
several values of $c_{i}$. Note that when $c_{i} = 1$, it represents the baseline method,
without the use of our proposed method. Before injecting bit-flips, weights are scaled by
$c_{i}$, and after injecting bit-flips, the weights are rescaled back to the original
scale by dividing by $c_{i}$. The absolute errors from bit-flips, $\lvert \frac{e(c_{i}
W_{i, j}, M_{i, j})}{c_{i}} \rvert$, are averaged across simulations and
presented in Fig. \ref{fig:err12}.

\begin{figure*}[htbp]
	\centering
	\begin{subfigure}{0.32\textwidth}
		\includegraphics[width=\textwidth]{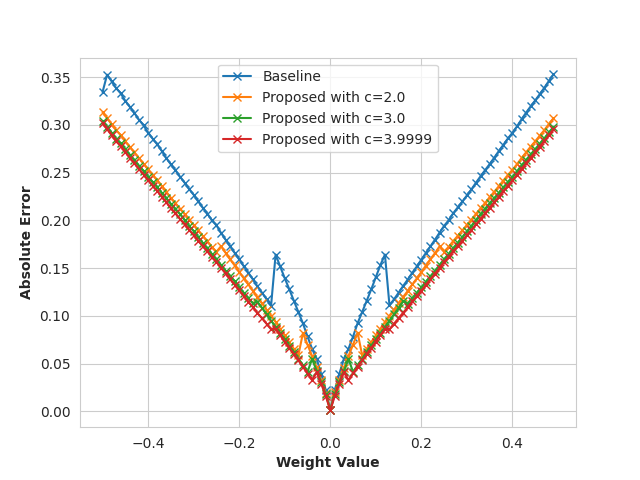}
		\caption{32-bit floating point}
		\label{fig:err12_1}
	\end{subfigure}
	\hfill
	\begin{subfigure}{0.32\textwidth}
		\includegraphics[width=\textwidth]{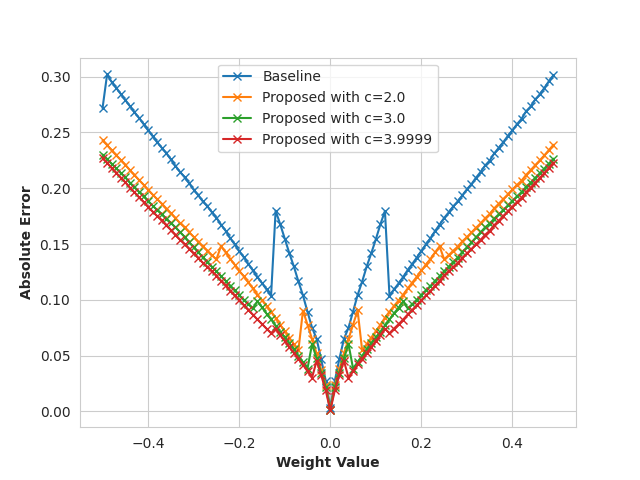}
		\caption{16-bit floating point}
		\label{fig:err12_2}
	\end{subfigure}
	\hfill
	\begin{subfigure}{0.32\textwidth}
		\includegraphics[width=\textwidth]{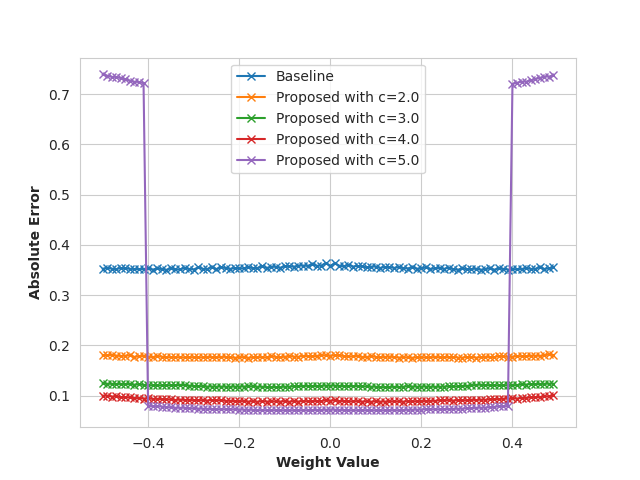}
		\caption{8-bit fixed point}
		\label{fig:err12_3}
	\end{subfigure}
	\caption{
        Absolute errors caused by bit-flips across three data types, comparing the
        baseline method and our proposed method.
	}
	\label{fig:err12}
\end{figure*}

From Fig. \ref{fig:err12}, higher values of the constant $c_{i}$ with our method result
in lower overall absolute errors compared to the baseline. This holds true for $c_{i}
W_{i, j} \in (-2, 2)$ for FP32 and FP16 and $c_{i}W_{i, j} \in [-2.0, 1.984375]$ for
Q2.5.

For Q2.5, when the scaled weights $c_{i}W_{i, j}$ exceed the Q2.5 data range of $[-2,
1.984375]$ \cite{fingeroff2010high}, overflow errors start to occur, as shown in
"Proposed with $c = 5.0$" in Fig. \ref{fig:err12_3}, where some of the scaled weights
fall out of the range of $[-2.0, 1.984375]$. On the other hand, for FP32 and FP16, when
the scaled weights $c_{i}W_{i, j}$ exceed the range of $(-2, 2)$ or when $c_{i} \geq 4$,
our assumption that $M_{i, j, 2} = 0$ alone is insufficient to protect the models from
drastic changes in the weights.

To further explain, when $c_{i}W_{i, j} < 2$, the exponent bits, or $B_{i, j, 2:9}$ are
arranged in the form $[0, X, X, ..., X]^{T}$, where $X$ indicates a "don't-care" term,
and $B_{i, j, a:b} = (B_{i, j, k})_{a \leq k \leq b}$. However, when $c_{i}W_{i, j} \geq
2$, $B_{i, j, 2:9} = [1, X, X, ..., X]^{T}$. When bit-flips occur with $c_{i}W_{i, j} <
2$, the only vulnerable bit position that can lead to significant error is $B_{i, j,
2}$, which is protected by our assumption. On the other hand, when $c_{i}W_{i, j} \geq
2$, these weights contain several vulnerable bit positions. For instance, if $c_{i}W_{i,
j} = 2$, bit-flips in $B_{i, j, 3}$, $B_{i, j, 4}$, and $B_{i, j, 5}$ can cause changes
greater than or equal to 131,072.

We provide two hypothesises for how $c_{i} > 1$ within given ranges helps reduce overall
absolute errors. The first hypothesis is that $e(c_{i} W_{i, j}, M_{i, j})$ grows more
slower than $c_{i}e(W_{i, j}, M_{i, j})$, meaning that dividing by $c_{i}$ reduces the
absolute error from bit-flips. The second hypothesis is that $e(c_{i} W_{i, j}, M_{i,
j})$ behaves similarly to additive noise, so dividing by $c_{i}$ reduces the absolute
error from bit-flips.

\subsection{Finding the Optimal $c_{i}$}

To find the optimal $c_{i}$, based on our analysis, the highest possible $c_{i}$ that
scales $W_{i}$ into the range  $(-2, 2)$ for FP32 and FP16, or $[-2, 1.984375]$ for
Q2.5,  must be determined. Since $W_{i}$ varies across layers, $t$ is defined as the
maximum value to which $W_{i}$ can be scaled. For FP32 and FP16, $t = 1.9999$ is
selected, and for Q2.5, $t = 1.97$ is chosen instead of $1.984375$ to avoid rounding
issues. The optimal $c_{i}$ is founded by \eqref{eq:c}.

\begin{equation}
	c_{i} = \frac{t}{\max(\lvert W_{i} \rvert)}
	\label{eq:c}
\end{equation}

\section{Experimental Results and Discussion}
In this paper, we conducted three experiments: fault injection with ResNet18 across
different BERs, fault injection across models, and a demonstration of how to reduce
the overall divisions by dividing output logits instead of weights.

\subsection{Fault Injection Across Different BERs}
In this experiment, we conducted fault injections with ResNet18 across different BERs
and $t$ values. Our ResNet18 weights were retrieved from \texttt{torchvision} package
\cite{torchvision2016}. For FP16 and Q2.5, the FP32 weights were converted to their
respective data types. All DNN operations were still performed using FP32 data type,
while the weights were simulated to experience bit-flips in their respective data types.
For all experiments with FP32 and FP16, we assumed that no bit-flips occur in the MSB
of the exponent term to ensure that the FP32 and FP16 models can operate under faults.

This experiment was designed to demonstrate how \( t \) influences the Top-1 Accuracy of
ResNet18 across different \( BER \) values. We performed fault injections with \(
BER \in \{10^{-3}, 10^{-4}, 10^{-5}\} \) and \( t \in \{0.5, 1.0, 1.5, 1.9999,
2.5, 3.0, 3.5, 4.0\} \) for FP32 and FP16. For Q2.5, we used the same \( BER \) values
but with \( t \in \{0.5, 1.0, 1.5, 1.97, 2.5, 3.0, 3.5, 4.0\} \). Monte Carlo
simulations were conducted to inject bit-flips into the weights, with 10 rounds for each
\( BER \) and \( t \).

The experimental results are visualized in Fig. \ref{fig:ber}, with error bars
indicating the standard deviation. The results are consistent with our analysis, showing
that our proposed method achieves the highest Top-1 Accuracy when $t = 1.9999$ for FP32
and FP16, and $t = 1.97$ for Q2.5. When $t > 1.9999$ for FP32 and FP16, and $t > 1.97$
for Q2.5, the Top-1 Accuracies drop significantly, and the error bars increase due to
more vulnerable bit positions for FP32 and FP16, as well as overflow errors for Q2.5.

\begin{figure*}[htbp]
	\centering
	\begin{subfigure}{0.32\textwidth}
		\includegraphics[width=\textwidth]{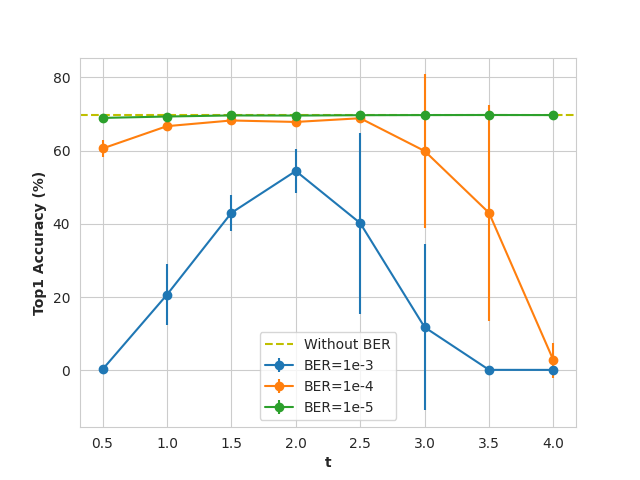}
		\caption{32-bit floating point}
		\label{fig:ber1}
	\end{subfigure}
	\hfill
	\begin{subfigure}{0.32\textwidth}
		\includegraphics[width=\textwidth]{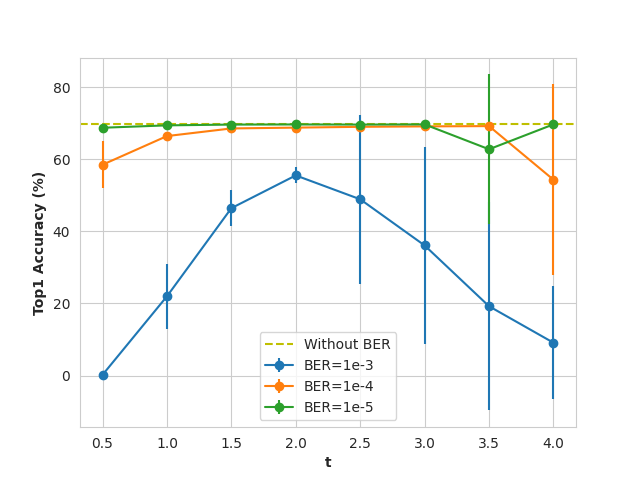}
		\caption{16-bit floating point}
		\label{fig:ber2}
	\end{subfigure}
	\hfill
	\begin{subfigure}{0.32\textwidth}
		\includegraphics[width=\textwidth]{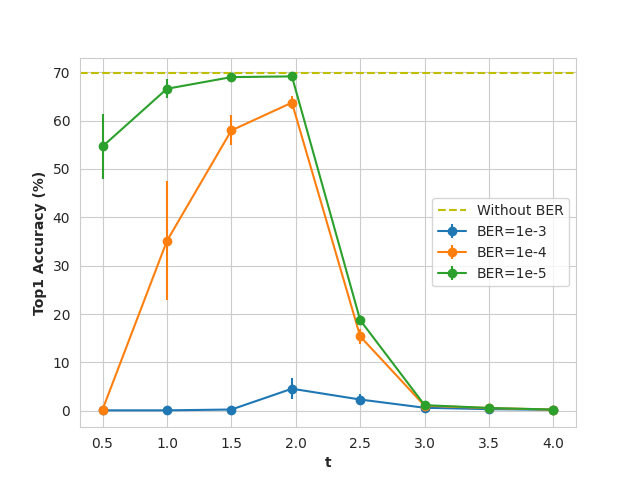}
		\caption{8-bit fixed point (Q2.5)}
		\label{fig:ber3}
	\end{subfigure}
    \caption{Top-1 Accuracy of ResNet18 under various $BER$ values and different $t$.}
	\label{fig:ber}
\end{figure*}

\subsection{Fault Injection Across Models}
In this experiment, we injected faults or bit-flips into ImageNet 2012
\cite{deng2009imagenet} pre-trained models: AlexNet \cite{krizhevsky2012imagenet},
ResNet18 \cite{he2016deep}, ResNet50 \cite{he2016deep}, and DenseNet169
\cite{huang2017densely}. These pre-trained weights were retrieved from the
\texttt{torchvision} package \cite{torchvision2016}. We performed 10 rounds of Monte
Carlo simulations to inject bit-flips into weights and reported the average Top-1
Accuracy with standard deviation, in format of Top-1 Accuracy $\pm$ Standard Deviation.
Before the experiments, the original Top-1 Accuracy of these models, or baseline scores
without bit-flips, were reported as shown in Table \ref{tab:origin}.

\begin{table}[htbp]
	\caption{
        The original Top-1 Accuracy of the models without bit-flips across the three
        data types.
	}
	\begin{center}
		\begin{tabular}{|c|c|c|c|}
			\hline
			\textbf{Model}               & \textbf{Datatype} & \textbf{Original Top1 Acc (\%)} \\
			\hline

			\multirow{2}{*}{AlexNet}     & FP32              & \textbf{56.55}                           \\
			\cline{2-3}
			                             & FP16              & \textbf{56.55}                           \\
			\cline{2-3}
			                             & Q2.5              & 51.24                           \\
			\hline

			\multirow{2}{*}{ResNet18}    & FP32              & \textbf{69.76}                           \\
			\cline{2-3}
			                             & FP16              & 69.75                           \\
			\cline{2-3}
			                             & Q2.5              & 66.87                           \\
			\hline

			\multirow{2}{*}{ResNet50}    & FP32              & \textbf{76.15}                           \\
			\cline{2-3}
			                             & FP16              & \textbf{76.15}                           \\
			\cline{2-3}
			                             & Q2.5              & 72.26                           \\
			\hline

			\multirow{2}{*}{DenseNet169} & FP32              & 75.59                           \\
			\cline{2-3}
			                             & FP16              & \textbf{75.6}                            \\
			\cline{2-3}
			                             & Q2.5              & 71.81                           \\
			\hline
		\end{tabular}
		\label{tab:origin}
	\end{center}
\end{table}

Monte Carlo simulations were performed with $BER = 10^{-3}$ for FP32 and FP16, and $BER
= 10^{-4}$ for Q2.5. A higher $BER$ was applied for FP32 and FP16 because higher model
failure rates were observed for Q2.5 at $BER = 10^{-3}$. Following the optimal $t$
values from the previous experiment, $t = 1.9999$ was used for FP32 and FP16, and $t
= 1.97$ for Q2.5. The experimental results are presented in Table \ref{tab:models}.

\begin{table}[htbp]
    \caption{Models under fault injection across three data types with different $t$ values.}
	\begin{center}
		\begin{tabular}{|c|c|c|c|}
			\hline
			\textbf{Model}               & \textbf{Datatype} & \textbf{t} & \textbf{Top1 Acc (\%)} \\
			\hline
			\multirow{2}{*}{AlexNet}     & FP32              & 1.0        & 2.68 $\pm$ 1.44       \\
			\cline{2-4}
			                             & FP32              & \textbf{1.9999}     &
			\textbf{34.26 $\pm$ 5.27}                                                            \\
			\hline
			\multirow{2}{*}{ResNet18}    & FP32              & 1.0        & 14.78 $\pm$
            7.04     \\
			\cline{2-4}
			                             & FP32              & \textbf{1.9999}     &
			\textbf{53.29 $\pm$ 4.62}                                                            \\
			\hline
			\multirow{2}{*}{ResNet50}    & FP32              & 1.0        & 4.83 $\pm$
            4.01      \\
			\cline{2-4}
			                             & FP32              & \textbf{1.9999}     &
			\textbf{53.78 $\pm$ 9.30}                                                             \\
			\hline
			\multirow{2}{*}{DenseNet169} & FP32              & 1.0        & 7.02 $\pm$
            4.45      \\
			\cline{2-4}
			                             & FP32              & \textbf{1.9999}     &
			\textbf{56.67 $\pm$ 4.67}                                                             \\
			\hline
			\hline

			\multirow{2}{*}{AlexNet}     & FP16              & 1.0        & 2.83 $\pm$ 1.27       \\

			\cline{2-4}
			                             & FP16              & \textbf{1.9999}     &
			\textbf{32.40 $\pm$ 4.72}                                                            \\
			\hline
			\multirow{2}{*}{ResNet18}    & FP16              & 1.0        & 23.48 $\pm$ 5.80     \\
			\cline{2-4}
			                             & FP16              & \textbf{1.9999}     &
			\textbf{56.34 $\pm$ 3.00}                                                             \\
			\hline
			\multirow{2}{*}{ResNet50}    & FP16              & 1.0        & 4.43 $\pm$ 4.33       \\
			\cline{2-4}
			                             & FP16              & \textbf{1.9999}     &
			\textbf{59.87 $\pm$ 4.62}                                                            \\
			\hline
			\multirow{2}{*}{DenseNet169} & FP16              & 1.0        & 3.85 $\pm$ 2.08      \\
			\cline{2-4}
			                             & FP16              & \textbf{1.9999}     &
			\textbf{58.59 $\pm$ 6.57}                                                            \\
			\hline
			\hline

			\multirow{2}{*}{AlexNet}     & Q2.5              & 1.0        & 6.35 $\pm$
            3.46      \\
			\cline{2-4}
			                             & Q2.5              & \textbf{1.97}       &
			\textbf{38.18 $\pm$ 4.35}                                                            \\
			\hline
			\multirow{2}{*}{ResNet18}    & Q2.5              & 1.0        & 32.59 $\pm$ 10.28    \\
			\cline{2-4}
			                             & Q2.5              & \textbf{1.97}       &
			\textbf{63.88 $\pm$ 2.20}                                                            \\
			\hline
			\multirow{2}{*}{ResNet50}    & Q2.5              & 1.0        & 14.38 $\pm$ 6.02      \\
			\cline{2-4}
			                             & Q2.5              & \textbf{1.97}       &
			\textbf{68.80 $\pm$ 2.55}                                                            \\
			\hline
			\multirow{2}{*}{DenseNet169} & Q2.5              & 1.0        & 16.02 $\pm$ 8.97     \\
			\cline{2-4}
			                             & Q2.5              & \textbf{1.97}       &
			\textbf{67.97 $\pm$ 3.45}                                                            \\
			\hline
		\end{tabular}
		\label{tab:models}
	\end{center}
\end{table}

From Table \ref{tab:models}, at $BER = 10^{-3}$, FP16 is overall more resilient to
bit-flips compared to FP32. We observe that our proposed method with the optimal $t$
improves the Top-1 Accuracy across all models and data types. On average, the
Top-1 Accuracy of models with bit-flips is improved by 42.17 for FP32, 43.15 for FP16,
and 42.37 for Q2.5 by simply multiplying and dividing weights with constants.

\subsection{Reducing Division Overhead}
Our proposed method requires dividing the weights $W_{i}$ by $c_{i}$ after reading from
fault-prone mediums. Since this division operation can be computationally expensive and
it must be applied to all $n$ weight parameters, we propose a method that reduces the
number of divisions by performing the divisions only at the output logits, as
illustrated in Fig. \ref{fig:multi-const}.

Instead of dividing the weights immediately after reading them from fault-prone mediums,
Fig. \ref{fig:multi-const} shows that DNN operations can be performed without division,
or with $W_{i}$ scaled by $c_{i}$. This method requires data types with a wide dynamic
range, such as floating point, particularly for operations involving higher magnitudes.

\begin{figure}[htbp]
	\centerline{\includegraphics[scale=0.4]{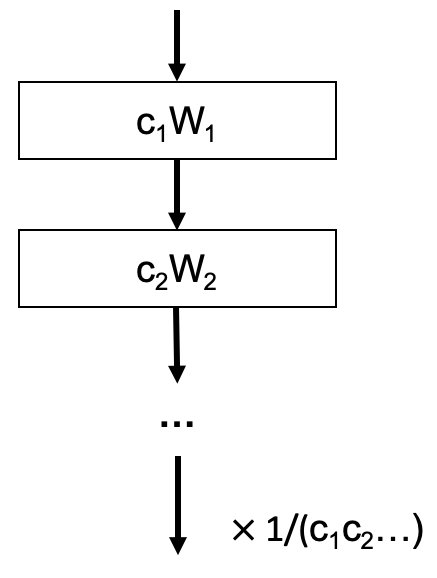}}
	\caption{
        Reduces the number of divisions by dividing the cumulative product of constants
        at the final logits, instead of dividing each weight and constant individually.
	}
	\label{fig:multi-const}
\end{figure}

This method requires $ab$ divisions, where $a$ is the number of classes and $b$ is the
batch size. Our method is advantageous when $a$ and $b$ are small, or when $ab < n$.
However, this method has two limitations. First, DNN operations must be performed at
a higher magnitude region. where the precision of floating-point arithmetic is
lower, potentially leading to a loss in precision. Second, while this method works
with models using ReLU activation, it is not suitable for models with highly non-linear
activation functions, such as Sigmoid or Tanh.

We conducted an experiment applying this method to FP32 AlexNet and obtained a Top-1
Accuracy of 54.13, compared to the original Top-1 Accuracy is 56.55 for FP32 AlexNet. We
attribute the loss in Top-1 Accuracy to the precision issues inherent in floating point
arithmetic.

\section{Conclusion}
In this paper, we propose a simple yet effective method to harden DNNs against bit-flips
by multiplying DNN weights by layer-wise constants before passing them to noisy mediums.
When using these weights, the weights with faults are divided by the same constants to
return them to their original scales. With this approach, we demonstrate that, within
certain ranges, the absolute errors from bit-flips can be reduced compared to the
baseline. Furthermore, we propose a method to reduce the overall number of divisions by
performing division only on the output logits, instead of on all weights.

\bibliographystyle{IEEEtran}
\bibliography{reference}
\end{document}